# The SAT Phase Transition[*]

Ke Xu  and  Wei Li

*National Laboratory of Software Development Environment*

*Department of Computer Science and Engineering*

*Beijing University of Aeronautics and Astronautics, Beijing 100083, P.R. China*

Email: kexu@nlsde.buaa.edu.cn

**Abstract**   Phase transition is an important feature of SAT problem. In this paper, for random $k$-SAT model, we prove that as $r$ (ratio of clauses to variables) increases, the structure of solutions will undergo a sudden change like satisfiability phase transition when $r$ reaches a threshold point ($r = r_{cr}$). This phenomenon shows that the satisfying truth assignments suddenly shift from being relatively different from each other to being very similar to each other.

The propositional satisfiability problem, or SAT problem for short, is a typical NP-complete problem. Designing fast algorithms to solve the SAT problem has not only important theoretical value, but also many immediate applications in areas such as formal development of software and logical inference engine. A lot of experimental studies suggest the following conjecture: for each $k$, there is some $r^*$ such that for each fixed value of $r < r^*$, random $k$-SAT with $n$ variables and $rn$ clauses is satisfiable with probability tending to 1 as $n \to +\infty$, and when $r > r^*$, unsatisfiable with probability tending to 1. We refer to this point as the crossover point or the phase transition point and this phenomenon is called the SAT phase transition. Another feature associated with the SAT phase transition is the hardness to solve SAT instances. It has been found in the experimental studies that most instances appear to be easy when they are far from phase transition area, while nearly all the algorithms exhibit poor performance on instances near phase transition point.

---

[*] Project supported by the National Natural Science Foundation of China (Grant No. 69433030)

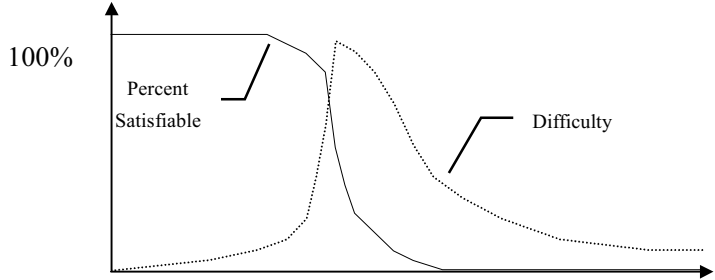

Fig. 1 Percent satisfiable and difficulty curves for SAT problem as a function of the ratio of clauses to variables.

Phase transition is an important feature of the SAT problem. Investigation on this phenomenon can help us to gain a better understanding of the SAT problem and design more efficient algorithms[3]. In the previous papers, several authors obtained the upper bounds for the phase transition point by studying the number of satisfying truth assignments. However, we can not give a complete analysis of the SAT phase transition if we only examine the number of solutions but do not consider the correlation among the satisfying truth assignments, because phase transition is not only a process of quantative change but also a process of qulitative change. Selman[4] and Clark[5] investigated the difficulty of solving SAT instances by using complete algorithms and incomplete algorithms respectively. It was found that although the SAT instances in the over-constrained area have a smaller number of solutions, search cost for instances with the same number of solutions tends to be less than at the phase transition. Their results show that the number of solutions is not the only factor determining problem hardness. Since the nature of search algorithms is to find solutions in the space of assignments, the structure of solutions will also have a close relation with the hardness of search. But there is still some lack of studies about how the structure of solutions varies with $r$ (ratio of clauses to variables). In this paper, we first define a parameter describing the extent to which the satisfying truth assignments are similar to each other, i.e., *major similarity degree* (definition 6). Then we prove that for random $k$-SAT model ($k \geq 5$), as $r$ increases, *major similarity degree* will undergo a sudden change like the satisfiability phase transition when $r$ reaches a threshold point ($r = r_{cr}$). So we refer to this phenomenon as the *phase transition of major similarity degree* ($s_{mj}$ *phase transition* for short). Paper [3] pointed out that other forms of phase transitions should also be studied besides the phase transitions of satisfiability[1] and search cost. However, so far we know little about proving the existence of some kind of phase transition theoretically for the general $k$-SAT model.

## 1 Definitions and lemmas

We first give some definitions related to the SAT problem before starting to analyze the structure of solutions.

**Definition 1** *Conjunctive normal forms and the SAT problem*

a) Let $U$ be a given set of Boolean variables, ranged over by $u_i$ where $1 \leq i \leq n$.

b) $t_i : U \to \{T, F\}$ is a truth assignment of $U$, where $1 \leq i \leq 2^n$.

c) If $u$ is a variable of $U$, then literals $u$ and $\neg u$ are positive and negative literals respectively. The literal is denoted by $L$.

d) A clause, denoted by $C$, is a set of literals combined only by connective $\vee$.

e) A formula in Conjunctive Normal Forms (CNF for short), denoted by $\Lambda$, is a set of clauses combined only by connective $\wedge$.

SAT problem is defined as follows: given a CNF formula $\Lambda$, the Satisfiability Problem (SAT for short) is to determine whether there exists a truth assignment that satisfies $\Lambda$. Random $k$-SAT formulae are obtained by choosing uniformly, independently and with replacement $m$ clauses from the space of clauses with $k$ distinct variables of $U$.

**Definition 2** *$\Phi$ stands for a function that maps a truth assignment into a point in the space*:

$$\Phi(t_i) = X_i(x_{i1}, x_{i2}, ..., x_{il}, ..., x_{in}), \text{ where } x_{il} = \begin{cases} 1, & t_i(u_l) = T, \\ 0, & t_i(u_l) = F. \end{cases}$$

Let $X_i(x_{i1}, x_{i2}, ..., x_{il}, ..., x_{in})$ denote a point in *n*-dimensional space, where $x_{il}$ is the $l$ th coordinate of this point ($1 \leq l \leq n$).

Each truth assignment can be uniquely mapped to a point in Euclidean space through the function $\Phi$. For example, there are $2^n$ truth assignments for a CNF formula defined on $U$ which correspond to $2^n$ points in *n*-dimensional space.

The correlation among truth assignments must be taken into consideration in order to study the structure of solutions. We first combine truth assignments into truth assignment pairs, and then introduce some new concepts to describe the relation between the two truth assignments in a truth assignment pair, such as *similarity number* and *similarity degree*.

---

[1] Recently, E. Friedgut made tremendous progress in proving the existence of the SAT phase transition.

**Definition 3**  *A truth assignment pair*

$<t_i, t_j>$ is a truth assignment pair of $U$ if and only if both $t_i$ and $t_j$ are two truth assignments of $U$. A truth assignment pair $<t_i, t_j>$ satisfies a CNF formula if and only if both $t_i$ and $t_j$ satisfy this formula. In this paper, all the truth assignments and truth assignment pairs are defined on $U$, and the set that consists of all the truth assignment pairs is denoted by $A_{pair}$.

**Note 1**.  $t_i$ may be equal to $t_j$, i.e., $<t_i, t_i>$ is a truth assignment pair. If $t_i \neq t_j$, $<t_i, t_j>$ and $<t_j, t_i>$ are two different truth assignment pairs.

Let $<t_i, t_j>$ be a truth assignment pair of $U$. By definition 2 and definition 3 we have:

$$\Phi(t_i) = X_i(x_{i1}, x_{i2}, ..., x_{il}, ..., x_{in}), \quad \Phi(t_j) = X_j(x_{j1}, x_{j2}, ..., x_{jl}, ..., x_{jn}).$$

**Definition 4**  *Similarity number*  $S^f : A_{pair} \to \{0, 1, 2, 3, ...\}$,

$$S^f(<t_i, t_j>) = \sum_{l=1}^{n} (1 - |x_{il} - x_{jl}|).$$

The *similarity number* of a truth assignment pair is equal to the number of variables at which the two truth assignments of this truth assignment pair take the identical values. By definition 4, it is obvious that $0 \leq S^f(<t_i, t_j>) = S \leq n$.

**Definition 5**  *Similarity degree*  $s^f : A_{pair} \to \mathrm{R}$,

$$s^f(<t_i, t_j>) = \frac{S^f(<t_i, t_j>)}{n}.$$

The *similarity degree* of a truth assignment pair determines the extent to which the two truth assignments in this truth assignment pair are similar to each other, i.e., the ratio of the *similarity number* to the total number of variables. The larger the value of $s^f(<t_i, t_j>)$, the more similar are the two truth assignments $t_i$ and $t_j$. By definition 5, it is obvious that

$0 \leq s^f(<t_i, t_j>) = s \leq 1$.

Let $\phi$ be a random $k$-SAT formula on $U$ with $rn$ clauses. $P(t_i)$ stands for the probability of truth assignment $t_i$ satisfying $\phi$, and $P(<t_i, t_j>)$ stands for the probability of $<t_i, t_j>$ satisfying $\phi$.

First, we know $P(t_i) = (1 - \frac{1}{2^k})^{rn}$. (1)

Given the total number of variables and the number of clauses, the probability of a truth assignment pair satisfying $\phi$ is only associated with the *similarity number*. The expression is as follows:

If $S^f(<t_i, t_j>) = S$, then $P(<t_i, t_j>) = \left( \dfrac{2^k - 2 + \dfrac{S(S-1)...(S-k+1)}{n(n-1)...(n-k+1)}}{2^k} \right)^{rn}$.

(2)

By definition 4, definition 5 and asymptotic analysis, we obtain the asymptotic estimate of $P(<t_i, t_j>)$ when $n$ approaches infinity:

If $s^f(<t_i, t_j>) = s$, then $P(<t_i, t_j>) = \sigma(s) e^{-nrg(s)}(1 + O(1/n))$ when $n \to +\infty$,

(3)

where

$$\sigma(s) = e^{r\rho(s)}, \quad \rho(s) = \frac{k(k-1)}{2(2^k - 2 + s^k)}(s^k - s^{k-1}), \quad g(s) = \ln 2^k - \ln(2^k - 2 + s^k).$$

Let $A_s$ be the set of truth assignment pairs whose *similarity degree* is equal to $s$.

The cardinality $|A_s| = 2^n C_n^{ns}$. (4)

We estimate the equation (4) by Stirling's formula:

$$|A_s| = \tau(s) e^{nh(s)}(1 + O(1/n)) \text{ when } n \to +\infty,$$ (5)

where the functions $\tau(s)$ and $h(s)$ are defined as:

$$\tau(s) = \begin{cases} 1 & (s = 0,1) \\ \dfrac{1}{\sqrt{2\pi ns(1-s)}} & (0 < s < 1) \end{cases},$$

$$h(s) = \begin{cases} \ln 2 & (s = 0,1) \\ \ln 2 - s \ln s - (1-s)\ln(1-s) & (0 < s < 1) \end{cases}.$$

Let $A_s^{Sat}$ denote the set of truth assignment pairs satisfying $\phi$ whose *similarity degree* is equal to $s$. It is obvious that the cardinality $|A_s^{Sat}|$ is a random variable. The expected value of this variable is denoted by $E(|A_s^{Sat}|)$.

$$E(|A_s^{Sat}|) = P(<t_i,t_j>)|A_s| = \varphi(s)e^{nf(s)}(1 + O(1/n)) \text{ when } n \to +\infty, \quad (6)$$

where $\varphi(s) = \sigma(s)\tau(s)$, $f(s) = h(s) - rg(s)$.

**Definition 6** *Major similarity degree*

Given $r$, if $s_0$ satisfies the following condition:

for every $0 \leq s \leq 1$, the inequality $\lim\limits_{n \to +\infty} \dfrac{\ln E(|A_s^{Sat}|) - \ln E(|A_{s_0}^{Sat}|)}{n} \leq 0$ holds. (7)

Then $s_0$ is *major similarity degree* that is denoted by $s_{mj}$ in this paper.

**Property 1** *Given $r$, if $s_0$ satisfies the following condition: for every $0 \leq s \leq 1$, there exists $M > 0$ such that $E(|A_s^{Sat}|) \leq E(|A_{s_0}^{Sat}|)$ whenever $n > M$. Then $s_0$ is major similarity degree.*

By definition 6 the proof is straightforward.

**Property 2** *Given $r$, if $s_0$ is not major similarity degree, then there exist $\delta > 0$ and $M > 0$ such that $E(|A_{s_{mj}}^{Sat}|) \geq e^{n\delta} E(|A_{s_0}^{Sat}|)$ whenever $n > M$.*

By definition 6 and equation (6) the poof can be easily obtained.

Property 1 and property 2 reveal that when $n$ approaches infinity the maximum points of $E(|A_s^{Sat}|)$ must be *major similarity degree*, and $E(|A_{s_{mj}}^{Sat}|)$ is $e^{n\delta}$ times larger than $E(|A_{s_0}^{Sat}|)$ (where $s_0$ is not *major similarity degree*). Therefore, *major similarity degree* describes the extent to which the satisfying truth assignments are similar to each other. The larger the values of *major similarity degree*, the more similar are the satisfying truth assignments.

**Lemma 1** $s_0$ is *major similarity degree* if and only if $s_0$ is a maximum point of $f(s)$.

By definition 6 and equation (6) the poof can be easily obtained.

By lemma 1 $s_0$ is *major similarity degree* if and only if $s_0$ is a maximum point of $f(s)$. The critical points of $f(s)$ satisfy the following equations:

$$f'(s) = h'(s) - rg'(s) = -\ln s + \ln(1-s) + r\frac{ks^{k-1}}{2^k - 2 + s^k} = 0, \tag{8}$$

$$\Leftrightarrow r(s) = \frac{h'(s)}{g'(s)} = \frac{1}{k}(\frac{2^k - 2}{s^{k-1}} + s)(\ln s - \ln(1-s)). \tag{9}$$

Equation (9) gives a functional relation between $r$ and the critical points. By examining the behaviour of this function, we can get the relation between $r$ and the maximum points, and so obtain the information about how *major similarity degree* behaves as $r$ varies. To investigate the behaviour of $r(s)$, we first analyze its derivatives.

**Lemma 2** *Given the clause length $k$ ($k \geq 5$), there exists one and only one root of equation $r''(s) = 0$ over $[0.5,1)$, denoted by $s_{02}$, and $r''(s) < 0$ on the interval $s < s_{02}$, and $r''(s) > 0$ on the interval $s > s_{02}$.*

**Proof.** The second derivative of $r(s)$ is as follows:

$$r''(s) = F(s)\frac{2^k - 2}{s^{k+1}(1-s)^2},$$

where

$$F_1(s) = k(k-1)(\ln s - \ln(1-s))(1-s)^2 - 2(k-1)(1-s) + 2s - 1,$$

$$F(s) = F_1(s) + \frac{s^k}{2^k - 2}, \quad F(0.5) < 0, \quad \lim_{s \to 1} F(s) > 0.$$

$F(s)$ is a continuous function over $[0.5,1)$. By the intermediate value theorem and the above equations, there exists at least one root $s_{02}$ such that $F(s_{02}) = 0$. We can further prove that there is at most one root. The proof is divided into the following two cases:

**Case 1** If there exists no root of equation $F_1'(s) = 0$ over $[0.5,1)$, it can be proved that $F'(s) > 0$, i.e., $F(s)$ is a strictly increasing function over $[0.5,1)$. Thus there is only one root of $F(s) = 0$.

**Case 2** If $F_1'(s_0) = 0$, then it can be deduced that $F_1(s_0) > 0$. In other words, $F_1(s)$ is greater than zero at the critical points. We can also prove that there exists only one root of $F(s) = 0$ over $[0.5,1)$.

Since the sign of $F(s)$ is the same as that of $r''(s)$, lemma 2 holds.

**Lemma 3** *Given the clause length $k$ ($k \geq 5$), there exist two and only two roots of equation $r'(s) = 0$ over $[0.5,1)$, denoted by $s_{01}$ and $s_{03}$ respectively (where $s_{01} < s_{02} < s_{03}$), and $r'(s) > 0$ on the interval $0.5 \leq s < s_{01}$, and $r'(s) < 0$ on the interval $s_{01} < s < s_{03}$, and $r'(s) > 0$ on the interval $s_{03} < s < 1$.*

**Proof.** $r'(s) \dfrac{ks^k}{(k-1)(2^k-2) - s^k} \leq -(\ln s - \ln(1-s)) + \dfrac{2^k - 1}{(k-1)(2^k-2) - 1} \dfrac{1}{(1-s)}$

Let $a = \dfrac{2^k - 1}{(k-1)(2^k - 2) - 1}$, $H(s) = \dfrac{a}{1-s} - (\ln s - \ln(1-s))$.

If $k \geq 5$, then $0.5 < \dfrac{1}{1+a} < 1$. It can be deduced that $H(\dfrac{1}{1+a}) < 0$, $r'(\dfrac{1}{1+a}) < 0$.

By lemma 2, $s_{02}$ is a minimum point of $r'(s)$. Therefore, if $k \geq 5$, then $r'(s_{02}) < 0$.

By the expression of $r'(s)$, it can be proved that $r'(0.5) > 0$, $\lim\limits_{s \to 1} r'(s) = +\infty$.

By lemma 2, $r'(s)$ is a strictly decreasing function over $[0.5, s_{02})$. Thus there exists $s_{01}$ such that $r'(s_{01}) = 0$, and $r'(s) > 0$ on the interval $0.5 \leq s < s_{01}$, and $r'(s) < 0$ on the interval $s_{01} < s \leq s_{02}$. By lemma 2, $r'(s)$ is a strictly increasing function over $(s_{02}, 1)$. Thus there exists $s_{03}$ such that $r'(s_{03}) = 0$, and $r'(s) < 0$ on the interval $s_{02} \leq s < s_{03}$, and $r'(s) > 0$ on the interval $s_{03} < s < 1$. Hence lemma 3 is proved.

By lemma 3 the behaviour of $r(s)$ as a function of $s$ can be easily obtained. $r(s)$ is a strictly increasing function on the intervals $0.5 \leq s \leq s_{01}$ and $s_{03} \leq s < 1$, and is a strictly decreasing function on the interval $s_{01} \leq s \leq s_{03}$. So we can define the inverse functions of $r(s)$ in every interval:

$$s_1(r) = r^{-1}(s): \quad [0, r(s_{01})] \to [0.5, s_{01}];$$

$$s_2(r) = r^{-1}(s): \quad [r(s_{03}), r(s_{01})] \to [s_{01}, s_{03}];$$

$$s_3(r) = r^{-1}(s): \quad [r(s_{03}), +\infty) \to [s_{03}, 1).$$

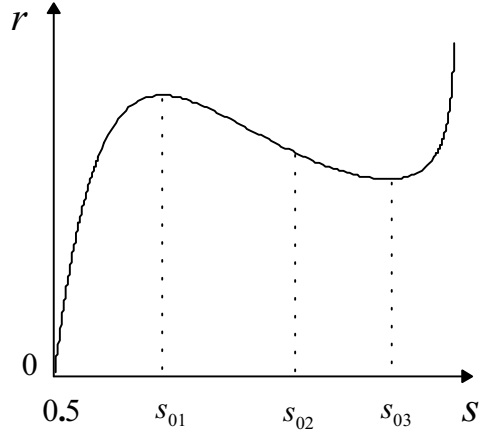

Fig. 2　The curve of $r(s)$ for random $k$-SAT model ($k \geq 5$)

We first analyze how *major similarity degree* varies with $r$ on the intervals $r < r(s_{03})$ and $r > r(s_{01})$. By equations (8) and (9), we have:

$$f'(s) = h'(s) - rg'(s) = g'(s)(\frac{h'(s)}{g'(s)} - r), \quad g'(s) = -\frac{ks^{k-1}}{2^k - 2 + s^k} < 0. \quad (10)$$

Given $r$, by fig. 2 and equation (10), if $r < r(s_{03})$, there exists only one critical point of $f(s)$, i.e. $s_1(r)$, and $f'(s) > 0$ over $[0.5, s_1(r))$, and $f'(s) < 0$ over $(s_1(r), 1)$. Therefore, $s_1(r)$ is the only one maximum point of $f(s)$. So by lemma 1, if $r < r(s_{03})$, then $s_1(r)$ is *major similarity degree*. Similarly, if $r > r(s_{01})$, then $s_3(r)$ is *major similarity degree*. $s_1(r)$ and $s_3(r)$ are strictly increasing functions. Therefore, if $r < r(s_{03})$ or $r > r(s_{01})$, *major similarity degree* will increase continuously as $r$ grows. However, it will be proved in the next section that the curve of *major similarity degree* as a function of $r$ is discontinuous at a threshold point on the interval $r(s_{03}) \leq r \leq r(s_{01})$.

## 2  Existence theorem of $s_{mj}$ phase transition

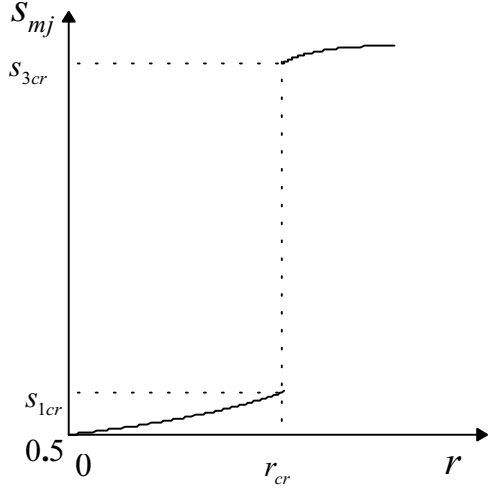

Fig. 3   The curve of $s_{mj}$ as a function of $r$ for random $k$-SAT model ($k \geq 5$)

**Note 2**   $s_{mj}$ phase transition occurs at $r = r_{cr}$

It will be proved in this section that as $r$ increases continuously over $[r(s_{03}), r(s_{01})]$, *major similarity degree* will undergo a sudden change like the SAT phase transition (see fig. 3). That is to say, there exists a threshold point $r_{cr}$ such that *major similarity degree* changes from a smaller value to a larger value abruptly when $r$ crosses this threshold point. In this paper, the phenomenon of this sudden change is called the *phase transition of major similarity degree* ($s_{mj}$ *phase transition* for short). The theorem and its proof are given below:

**Theorem 1**   *For random $k$-SAT model, given the clause length $k$ ($k \geq 5$), there exists a threshold point $r_{cr}$ (the value of $r_{cr}$ varies with $k$) such that for any small positive value $\varepsilon$, if $r = r_{cr} - \varepsilon$, then $s_{mj} < s_{1cr}$ and $\lim_{r \to r_{cr}-0} s_{mj} = s_{1cr}$; if $r = r_{cr} + \varepsilon$, then $s_{mj} > s_{3cr}$ and $\lim_{r \to r_{cr}+0} s_{mj} = s_{3cr}$, where $s_{1cr} < s_{3cr}$.*

**Proof.**   Given $r$, if $r(s_{03}) < r < r(s_{01})$, there are three critical points of $f(s)$ which are $s_1(r)$, $s_2(r)$ and $s_3(r)$. Similarly, by equation (10) it can be easily deduced that $s_1(r)$ and $s_3(r)$ are local maximum points and $s_2(r)$ is a local minimum point. *Major similarity degree* can be easily obtained by deciding which local maximum value is greater. We first define the following function:

$$F(r) = f(s_1(r)) - f(s_3(r)), \quad r(s_{03}) \leq r \leq r(s_{01}). \tag{11}$$

By fig. 2 and equation (10), if $r = r(s_{03})$, then $s_1(r(s_{03}))$ is the only one maximum point of $f(s)$. Hence we have:

$$F(r(s_{03})) = f(s_1(r(s_{03}))) - f(s_3(r(s_{03}))) > 0. \tag{12}$$

Similarly, if $r = r(s_{01})$, then $s_3(r(s_{01}))$ is the only one maximum point of $f(s)$. Thus we deduce:

$$F(r(s_{01})) = f(s_1(r(s_{01}))) - f(s_3(r(s_{01}))) < 0. \tag{13}$$

The first derivative of $F(r)$ is as follows:

$$F'(r) = h'(s_1(r))s_1'(r) - g(s_1(r)) - rg'(s_1(r))s_1'(r) \\ - h'(s_3(r))s_3'(r) + g(s_3(r)) + rg'(s_3(r))s_3'(r). \tag{14}$$

By use of the condition that $s_1(r)$ and $s_3(r)$ are the critical points of $f(s)$, we have:

$$h'(s_1(r)) - rg'(s_1(r)) = 0, \quad h'(s_3(r)) - rg'(s_3(r)) = 0. \tag{15}$$

Substituting equation (15) into equation (14), we obtain:

$$F'(r) = \ln(2^k - 2 + (s_1(r))^k) - \ln(2^k - 2 + (s_3(r))^k). \tag{16}$$

It is obvious that $s_1(r) \leq s_{01} < s_{03} \leq s_3(r)$. Thus $F'(r) < 0$. \hfill (17)

By the intermediate value theorem and the equations (12), (13) and (17), there exists only one root $r_{cr}$ of $F(r) = 0$, and the following facts hold:

If $r(s_{03}) \leq r < r_{cr}$, then $f(s_1(r)) > f(s_3(r))$. Hence $s_{mj} = s_1(r) < s_1(r_{cr})$.

If $r_{cr} < r \leq r(s_{01})$, then $f(s_1(r)) < f(s_3(r))$. Hence $s_{mj} = s_3(r) > s_3(r_{cr})$.

Let $s_{1cr} = s_1(r_{cr})$ and $s_{3cr} = s_3(r_{cr})$. It is obvious that $s_{1cr} < s_{3cr}$.

If $r = r_{cr}$, then $f(s_{1cr}) = f(s_{3cr})$. Both $s_{1cr}$ and $s_{3cr}$ are maximum points. Thus $s_{mj} = s_{1cr}, s_{3cr}$.

Since $s_1(r)$, $s_3(r)$ are continuous and strictly increasing functions, we have:

If $r = r_{cr} - \varepsilon$, then $s_{mj} < s_{1cr}$ and $\lim_{r \to r_{cr}-0} s_{mj} = s_{1cr}$.

If $r = r_{cr} + \varepsilon$, then $s_{mj} > s_{3cr}$ and $\lim_{r \to r_{cr}+0} s_{mj} = s_{3cr}$.

Therefore, theorem 1 holds.

## 3  Conclusion

Theorem 1 indicates that a phase transition phenomenon does occur for *major similarity degree*, which changes from a smaller value $s_{1cr}$ to a larger value $s_{3cr}$ abruptly when $r$ crosses the threshold point $r = r_{cr}$. By definition 6 in this paper, *major similarity degree* is a parameter describing the structure of solutions whose values show how much the satisfying truth assignments are similar to each other. The larger the values of *major similarity degree*, the more similar are the satisfying truth assignments. Therefore, this phenomenon of phase transition can be described as follows: as $r$ (ratio of clauses to variables) increases, the structure of solutions will undergo a phase transition phenomenon when $r$ reaches a threshold point ($r = r_{cr}$), which shows that the satisfying truth assignments suddenly shift from being relatively different from each other to being very similar to each other.

This paper proves that there really exist other forms of phase transitions for the SAT problem besides the satisfiability phase transition. This is analogous to many kinds of phase transitions in nature, i.e., there will always occur some other phenomena of sudden changes as some kind of phase transition takes place. Although these phenomena behave in different forms, there often exist deep connections among them that reflect the nature of the phase transition from different aspects. Therefore, we should study the SAT phase transition from various angles in order to get a better understanding of the SAT problem. Since we still know little about proving the existence of some kind of phase transition theoretically for the general $k$-SAT model, this paper may give some new insight into this field.

**Acknowledgement**   During the preparation of this paper, there were many helpful talks between the author and other researchers, such as Liang Dongmin, Luan Shangmin and Huang Xiong.